\documentclass[10pt,letterpaper]{article}
\usepackage[title,titletoc]{appendix}

\usepackage[numbers]{natbib}  
\usepackage{graphicx}
\usepackage{amsmath}
\usepackage{textcomp}
\usepackage{csquotes}
\usepackage{placeins}
\usepackage{multirow}
\usepackage{booktabs}
\usepackage{float}
\usepackage{xcolor}
\usepackage[colorlinks=true,linkcolor=blue,citecolor=blue]{hyperref}

\usepackage{array}
\usepackage{tabularx}

\usepackage{authblk}

\title{Precise Length Control in Large Language Models}
\author{Bradley Butcher, Michael O'Keefe$^1$, James Titchener$^{1,}$\thanks{Corresponding author: james.titchener@raytheon.co.uk}\vspace{1em}\\\normalsize $^1$Strategic Research Group, Raytheon UK}

\date{}  
\begin{document}
\maketitle

\begin{abstract}
    Large Language Models (LLMs) are increasingly used in production systems, powering applications such as chatbots, summarization, and question answering. 
    Despite their success, controlling the length of their response remains a significant challenge, 
    particularly for tasks requiring structured outputs or specific levels of detail. 
    In this work, we propose a method to adapt pre-trained decoder-only LLMs for precise control of response length.
    Our approach incorporates a secondary length-difference positional encoding (LDPE) into the input embeddings, which counts down to a user-set response termination length.
    Fine-tuning with LDPE allows the model to learn to terminate responses coherently at the desired length, achieving mean token errors of less than 3 tokens. 
    We also introduce Max New Tokens++, an extension that enables flexible upper-bound length control, rather than an exact target. 
    Experimental results on tasks such as question answering and document summarization demonstrate that our method enables precise length control without compromising response quality.
\end{abstract}

\section{Introduction}
\label{sec:intro}

In recent years, Large Language Models (LLMs) have become essential components of various production systems,
revolutionising the way we interact with technology.
From chatbots like ChatGPT to applications in document summarisation,
question answering, and content rewriting, LLMs have demonstrated remarkable capabilities in understanding and generating human-like text \cite{devlin2018bert, brown2020language}.
These models, trained on vast amounts of data,
have the potential to transform industries and enhance user experiences across a wide range of domains.

However, despite their impressive performance,
LLMs often generate unpredictable length responses even when prompted with specific instructions for how long the response should be.
This is particularly problematic in tasks that require structured outputs, or varying levels of detail,
such as document summarisation~\cite{gambhir2017recent}.
In situations where concise summaries are needed, an LLM might generate overly lengthy responses,
while in cases where more comprehensive summaries are desired, the model might produce insufficient detail.
This lack of control over response length limits the practical applicability of LLMs in real-world scenarios.

To address this limitation, we present a novel method to fine-tune pre-trained LLMs to enable precise control of response length.
Our approach adapts the length-difference positional encoding (LDPE) proposed by \cite{takase2019positional} to decoder-only transformer architectures.
This is achieved by incorporating a countdown mechanism that starts from a predetermined response length.
During the training process the model considers this countdown as a signal and implicitly learns the concept of \enquote{token budget},
i.e. how many tokens are remaining to generate an appropriate response.
A user-specified token budget can then be provided at inference time,
thereby facilitating the generation of responses with specified lengths for downstream tasks.

The main contributions of our paper are as follows:
\begin{enumerate}
	\item We adapt LDPE for length control in decoder-only LLMs and investigate the
	optimal way to add the encodings to a question response pair.
	\item We demonstrate the effectiveness of our approach through experiments on various tasks, including question answering and document summarisation.
	\item Our results demonstrate precise control of response length without reduction in response quality, enhancing the flexibility and applicability of LLMs in real-world scenarios.
\end{enumerate}

The remainder of this paper is organised as follows.
Section~\ref{sec:related} provides an overview of related work in the field of LLMs and output length control.
Section~\ref{sec:method} describes our methodology, explaining how we fine-tune LLMs with an additional positional encoding to provide a countdown mechanism.
Section~\ref{sec:experiments} details our experimental setup and the tasks used for evaluation.
Section~\ref{sec:results} presents and discusses the results obtained from our experiments.
Section~\ref{sec:conclusion} concludes the paper and outlines potential future directions for research.
Finally, Section~\ref{sec:limitations} details some of the limitations of the work which could be addressed to improve the method.
\section{Related Work}
\label{sec:related}



Positional encoding is used to directly provide transformers with
information about the absolute or relative position of each token \cite{vaswani2017attention,shaw2018self}.
Multiple techniques have been proposed to augment the standard positional encoding
with a secondary encoding containing useful positional information.
For example using a secondary encoding to align digit positions of operands has been shown to provide
greatly improved numerical calculation capabilities \cite{mcleish2024transformers}.
More generally, an additional learned context dependent encoding allows models to perform a
variety of tasks requiring different levels of positional understanding \cite{golovneva2024contextual}.
Here we explore adding a secondary positional encoding scheme to allow direct control over the response length.

Previous approaches for length control in text generation have primarily focused on encoder-decoder models for tasks like summarisation.
For instance, \cite{kikuchi2016controlling} proposed methods to initialise the LSTM cell of the decoder with a learnable vector based on the desired length
(LenInit) or input an embedding representing the remaining length at each decoding step (LenEmb).
This work was extended in~\cite{DBLP:journals/corr/abs-2106-00316} (LenAtten) to separate the length information from the decoder hidden states to better exploit the remaining length information.
In \cite{liu2022length} a length aware attention mechanism is used to modifiy the attention matrix based on the length-budget dynamically during the decoding process.
Most relevant to our work, \cite{takase2019positional} introduced length-difference positional encoding (LDPE),
which modifies the sinusoidal positional encoding in encoder-decoder transformer models to give information about the remaining length to the end of the text.

While effective for custom encoder-decoder architectures, these methods are not compatible with the large pre-trained decoder-only LLMs that have become prevalent in recent years.
Decoder-only LLMs like GPT-3 \cite{brown2020language} do not have an explicit length input during generation, making length control more challenging.
It has been shown that the response length is important for human evaluation scores, and early work
suggested controlling the response length by simply restricting the generation of the end-of-text token until a predefined response length was met \cite{roller2021recipes}.
However, this provides only coarse-grained control and may lead to unnatural responses.

Prompt-based fine-tuning is the most common method used to control the response length in LLMs \cite{Zhang2022LatentPT,Liu2021PretrainPA}.
For example \cite{goyal2022news} prompts GPT-3 to summarise articles in a given number of sentences.
In~\cite{Jie2023PromptBasedLC} reinforcement learning is used to fine-tune decoder-only LLMs using a rule-based reward model.
The type of prompt (more than, less than, equal to, between) determines the reward function,
which is calculated using the target and output length.
Although these methods are applicable to pre-trained decoder-only LLMs and provide some high level control over the response length,
they do not provide precise control of the response length down to the word or token level.
In this work we demonstrate a simple method to achieve precise token level control of the response length that can be applied directly to pre-trained LLMs.
\section{Method}
\label{sec:method}
In this section, we describe our proposed method for controlling the output length of LLMs by adapting the length-difference positional encoding (LDPE) proposed by \cite{takase2019positional} to the decoder-only transformer architecture.

Our approach consists of two main components:
(1) fine-tuning pre-trained LLMs with an added encoding counting down to the response termination point, and
(2) applying the reverse encoding during inference to control the length of the generated response.

\subsection{Adding Reverse Positional Encodings to SoTA LLMs}
In the transformer architecture \cite{vaswani2017attention}, positional encodings are used to provide information about the position of each token in the input sequence.
The LDPE scheme, as introduced by \cite{takase2019positional}, essentially reverses these encodings to show the model the remaining distance to the end of the generated sequence.
In this paper we demonstrate an effective way to integrate a similar length control scheme into current state-of-the-art decoder-only LLMs.

Current open source transformer models have typically undergone extensive pre-training using rotary positional embeddings (RoPE) to provide positional information \cite{SU2024}.
Because of this it would likely be detrimental to the model's performance to modify the RoPE to apply length control via LDPE.
Thus in order to avoid interfering with the RoPE, we apply LDPE by adding an additional absolute positional encoding to the input token embeddings.

\begin{figure*}[h]
	\centering
	\includegraphics[width=0.99\textwidth]{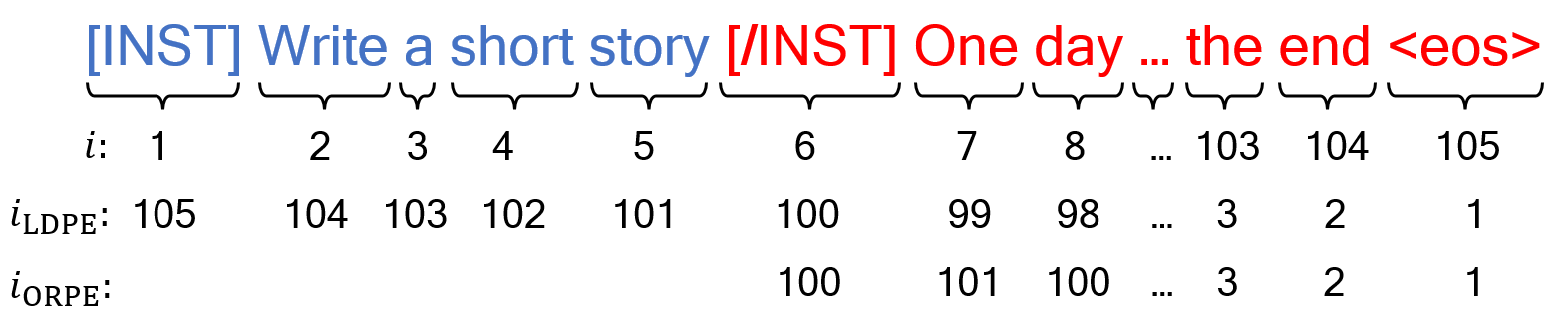}
	\caption{
		A simplified diagram of the reverse positional encoding methods applied a target response length
		of 100 tokens and a question length of 5 tokens ($L=105$ and $n=5$).
		For LDPE a positional encoding of $PE{(i_{\text{LDPE}},k)}$ is added to each token embedding.
		For ORPE no encoding is added to the prompt part of the text, then a encoding countdown from
		$i_{\text{ORPE}} = L-n$ to $i_{\text{ORPE}} = 1$ is added to the response.
	}
	\label{fig:orpe}
\end{figure*}

An additional consideration when applying LDPE to decoder-only models is that, because they process both the user's prompt and the model's response as a
 continuous sequence without distinct encoding and decoding stages, they lack a clear separation between the two.
This introduces some ambiguity about whether the reverse encodings should start at the beginning of the prompt, or only to the beginning of the models response.
We investigate applying the LDPE encoding to the entire prompt-response pair, and also only applying it to the models response.
We refer to the former simply as LDPE, and the latter as offset reverse positional encoding (ORPE), since the encoding is effectively offset by the prompt length (see Fig. \ref{fig:orpe}).

To represent the reverse positional encodings in the embedding space of a LLM we use the
standard sinusoidal encodings \cite{takase2019positional,vaswani2017attention}:

\begin{equation}
	PE{(i, 2k)} = \sin \left( \frac{i}{10000^{\frac{2k}{d}}} \right),
	\label{PE_1}
\end{equation}

\begin{equation}
	PE{(i, 2k+1)} = \cos \left( \frac{i}{10000^{\frac{2k}{d}}} \right).
	\label{PE_2}
\end{equation}

Here $i$ represents the tokens position in the sequence, $k$ is the index within the embedding dimension, and $d$ is the total number of dimensions in the embedding space.

In order to utilise the LDPE we adjust the order that the positional encodings are used by reversing them to count down towards $i = 1$.
Additionally for the case of Offset Reverse Positional Encoding (ORPE) we introduce an offset so that no encoding is applied to the prompt, and the countdown starts from the end of the input prompt.

Formally, let $(X,Y) = {x_1, \dots, x_n, y_{n+1}, \dots y_{L}}$ be a sequence of tokens consisting of a
prompt and response, where $n$ is the prompt length and $L$ is the total length of the prompt response pair.
For LDPE we augment the $i^{th}$ token embedding by adding the corresponding LDPE encoding $PE{(i_{\text{LDPE}},k)}$,
where, $i_{\text{LDPE}} = (L + 1) - i$.

Similarly, for ORPE we add the encoding $PE{(i_{\text{ORPE}},k)}$, where $i_{\text{ORPE}} = (L + 1) - i$, but no encoding is added to the tokens that make up the prompt (i.e. for $i \leq n$).
By offsetting the start of the reverse positional encoding to coincide with the beginning of the model's response,
we ensure that the countdown only applies to the generated text, not the input prompt.
The arrangement of the reverse positional encodings is demonstrated in Figure~\ref{fig:orpe},
using word-level tokenization for illustrative purposes only.






\subsection{Fine-tuning with Reverse Positional Encodings}
To incorporate the additional encoding into the LLM, we fine-tune pre-trained open-source LLMs with the LDPE or ORPE embeddings added to the original input embeddings.

Given a fine-tuning dataset of prompt-response pairs,
\begin{equation}
\mathcal{D} = {(X_1, Y_1), (X_2, Y_2), \dots, (X_m, Y_m)}
\end{equation}

where $X_i$ is the prompt and $Y_i$ is the corresponding target response, we fine-tune the LLM to minimise the following loss function:
\begin{equation}
	\mathcal{L} = -\sum_{i=1}^m \sum_{j=1}^{|Y_i|} \log P(y_{i,j} | X_i, R_{i,<j}, y_{i,<j})
\end{equation}

where $y_{i,j}$ is the $j$-th token of the target response $Y_i$, and $y_{i,<j}$ denotes the tokens preceding $y_{i,j}$ with corresponding LDPE (or ORPE) encodings $R_{i,<j}$.
By adding the reverse positional encodings during fine-tuning, the LLM learns to generate responses of the desired length,
as the model can use the countdown information provided by the encodings.

\subsubsection{Adapting Existing LLMs}

Directly adding reverse positional encodings to the token embeddings of an off-the-shelf LLM may not yield optimal results due to differences in scale between the learned embeddings and the positional encodings. To address this, we introduce a scaling term that balances the magnitude of ORPE with the token embeddings.
Let $E = {e_1, e_2, \dots, e_n}$ be the token embeddings of the input sequence $X$. We compute the scaled embeddings for both LDPE and ORPE ($R'$) as follows:
\begin{equation}
	R' = R \cdot \frac{|E|_F}{|R|_F}
	\label{eq:scaled_orpe}
\end{equation}
where $|E|_F$ and $|R|_F$ denote the Frobenius norm of the token embeddings and the reverse positional embeddings, respectively.
By scaling the reverse positional embeddings with the ratio of the Frobenius norms, we ensure that they have a similar magnitude to the token embeddings, allowing the LLM to effectively incorporate the length control signal during fine-tuning and inference.

The final input to the LLM, $\hat{E}$, is computed by adding the scaled reverse positional embeddings to the token embeddings:
\begin{equation}
	\hat{E} = E + R'
\end{equation}

This adaptation allows existing LLMs to benefit from the length-controlling properties of reverse positional embeddings without requiring significant modifications to the model architecture.

\subsection{Inference with Length Control}
At inference time, we control the length of the generated response by providing the a reverse positional embedding that counts down to the target length.
Given a prompt $X$ and a desired response length $L$,
we generate the reverse positional encoding $R$ and normalize it according to Equation~\ref{eq:scaled_orpe} before adding $R$ to the token embeddings.
The LLM then generates the response token by token, conditioning on the prompt, $R$, and the previously generated tokens.
The generation process continues until an end-of-sequence token is generated,
which given the fine-tuning process should be approximately around the target length.
By manipulating the reverse positional embeddings termination length, we can control the length of the generated response, enabling the LLM to produce concise or verbose answers as required by the application.

\subsection{Max New Tokens++}
While generating responses with exact lengths can be useful in certain applications,
there are scenarios where specifying upper-bound on the desired length is more appropriate.
The widely-used transformers library \cite{wolf2019huggingface} achieves this simply by stopping generation once the sequence exceeds the user specified maximum length.
However, this approach does not provide the model with an awareness of its remaining
\enquote{token budget} during the generation process, which may lead to suboptimal response quality.

To address this limitation, we propose a content-aware length control method that allows
the model to learn to terminate the response at or before a user set maximum response length.
During fine-tuning with the LDPE (or ORPE) reverse encodings, we introduce a random increase in the length of the LDPE sequence.
This has the effect of exposing the model to training samples where the ground truth text terminates naturally before the final token of the LDPE.
The random increase in length is sampled from a  half-normal distribution.
The long tail of the half-normal distribution is truncated to a user set maximum shift.


The random positive shift is added to the position index of reverse positional encoding.
Thus the $i^{th}$ encoding is given by ${PE}(i_{\text{shift}},k)$,
where $i_{\text{shift}}$ is determined randomly for each training sample according to,

\begin{equation}
	\label{eq:LDPE+shift2}
	i_{\text{shift}} = (L + 1) - i + \text{HalfNormal}(\sigma)
\end{equation}


By exposing the model to various target response lengths for the same input prompt during training,
we encourage it to learn to generate coherent and relevant responses that can be equal to or shorter than the user set response length.
The half-normal distribution is chosen because it has a higher probability of yielding small values or the shift,
ensuring that the model still observes the original target lengths more frequently,
providing a valuable learning signal.

To further improve the model's ability to generate responses with varying lengths, we employ a curriculum learning approach \cite{bengio2009curriculum} by gradually increasing the scale parameter $\sigma$ of the half-normal distribution throughout the training process. Initially, $\sigma$ is set to a small value, resulting in shifts that are close to zero. This encourages the model to focus on learning to generate responses that closely match the true response length. As training progresses, $\sigma$ is slowly increased according to an exponential schedule:
\begin{equation}
	\label{eq:exponential_schedule}
	\sigma_t = \sigma_0 \exp \left(\frac{t}{T} \log \left(\frac{\sigma_\text{max}}{\sigma_0}\right)\right)
\end{equation}
where $\sigma_0$ and $\sigma_\text{max}$ are the initial and maximum values of $\sigma$, respectively, $t$ is the current training step, and $T$ is the total number of training steps. The values of $\sigma_0$ and $\sigma_\text{max}$ are treated as hyperparameters and can be tuned to control the level of length variability and the trade-off between conciseness and informativeness.

Our updated \textit{Max New Tokens++} method offers a principled approach to generating responses with a desired maximum length while allowing the model to learn when to terminate the response based on the input prompt and generated content. This enables the model to produce more concise and relevant responses compared to the naive generation stopping criterion, enhancing the flexibility and applicability of language models in real-world scenarios.
\section{Experimental Setup}
\label{sec:experiments}
This section provides an overview of the models,
data and hyperparameters used to evaluate the performance of the LDPE and ORPE methods.
For brevity, in the following section, only the LDPE results will be shown and discussed,
and the corresponding ORPE results can be found in the Appendix.

\paragraph{Data}
A combination of the OpenOrca~\cite{OpenOrca} and MMLU~\cite{hendrycks2021measuring} datasets was used for training.
A training set of 110,000 samples was constructed by combining 100,000 OpenOrca samples with 10,000 MMLU samples (from across all topics).
These datasets were combined to cover a wider range of topics and sequence lengths,
and to utilise both human and synthetically generated data.
An evaluation set of 200 samples was constructed in the same manner,
taking 100 samples from each dataset.
All training and evaluation data was in English,
including the datasets and benchmarks discussed in Sections~\ref{sec:summarisation} and ~\ref{sec:benchmarking}.

\paragraph{Models}
Experiments were performed by fine-tuning both Mistral 7B~\cite{jiang2023mistral} and Llama3 8B~\cite{llama3}.
These models were chosen as they are exemplary representations of modern LLMs with robust architectures and significant parameter counts,
whilst still being small enough to train on a single NVIDIA RTX A6000 GPU.
The instruct versions of both models were used, and their forward pass methods were modified to add the appropriate LDPE encodings.

\paragraph{Baseline Models}
As a baseline for general model response quality we simply used the pretrained Mistral 7B Instruct and Llama3 8B Instruct models without any reverse encoding added.
Additionally as a baseline method of length control we fine-tuned Mistral with the target response length included in the prompt, for example \enquote{Answer the following question in 112 tokens:}.

\paragraph{Hyperparameters}
Both models were trained for a single epoch using the entire training dataset and a batch size of 1 with gradient accumulation of 5.
All models were trained using the AdamW optimiser~\cite{loshchilov2019decoupled} with $\beta_1 = 0.9$, $\beta_2 = 0.99$ and $\epsilon = 1 \times 10^{-8}$,
with a learning rate of 0.0003 and a linear learning rate schedule.
These hyperparameter settings were obtained via manual experimentation and evaluation until satisfactory results were obtained.

\paragraph{Low Rank Adaptation}
Low Rank Adaptation (LoRA)~\cite{DBLP:journals/corr/abs-2106-09685} was employed to efficiently train all models.
LoRA allows for efficient incorporation of the additional reverse positional encodings without overwriting the pretraining weights.
A rank of 16, an $\alpha$ of 32 and a dropout rate of 0.05 was used to train all models.
\section{Results}
\label{sec:results}
\subsection{Exact Length Control}
\subsubsection{Q\&A}
\label{sec:results_qa}
\begin{figure*}[h!]
	\centering
	\includegraphics[width=0.49\textwidth]{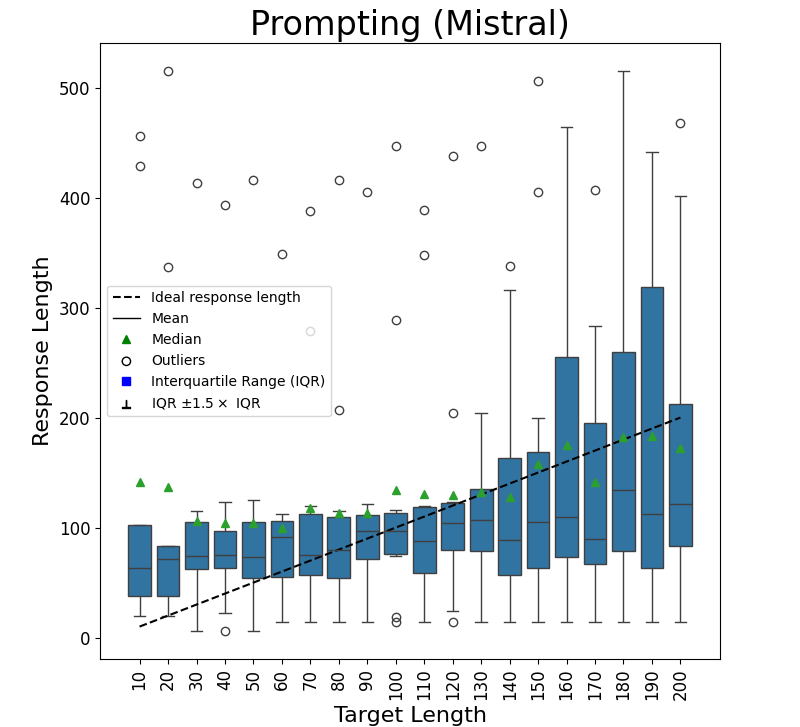}
	\includegraphics[width=0.49\textwidth]{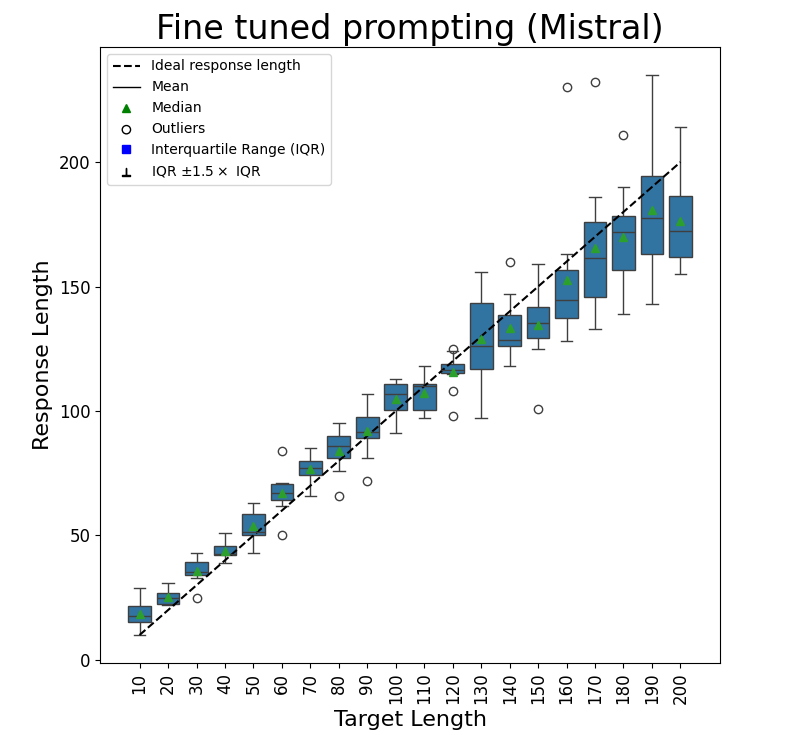}
	\includegraphics[width=0.49\textwidth]{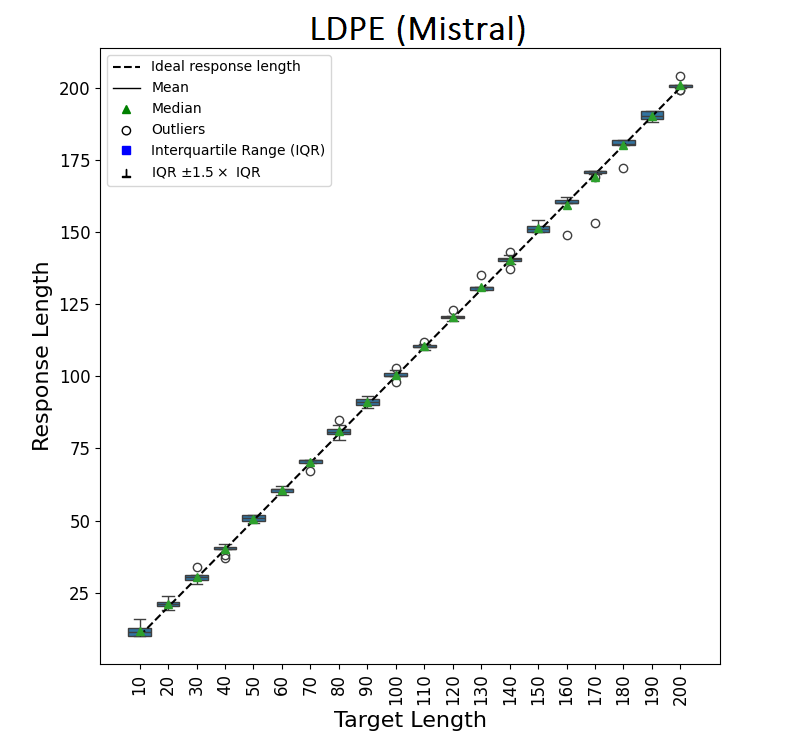}
	\includegraphics[width=0.49\textwidth]{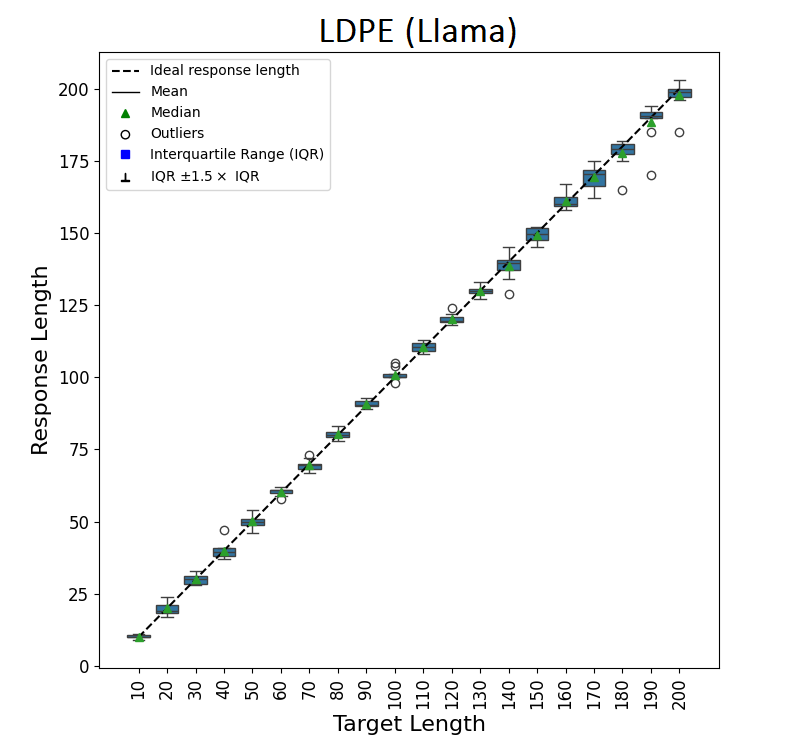}
	\caption{
		Comparison of target and response lengths for different length control approaches on a question-answering task.
		The ideal response length is indicated by the dashed line in each panel.
		\textbf{Top left:} results for prompting Mistral without length control.
		\textbf{Top right:} results for fine-tuned prompting Mistral.
		\textbf{Bottom left:} results for LDPE fine tuned Mistral 7B model.
		\textbf{Bottom right:} results for LDPE fine tuned Llama3 8B model.
	}
	\label{fig:qa_length_comparison}
\end{figure*}

To demonstrate the effectiveness of the method we evaluated its performance on the OpenOrca dataset.
A set of 20 samples were selected and different length responses were generated for each,
ranging from 10 to 200 tokens.
The generated responses length was then compared to the target length.
This analysis was performed for three different approaches:
prompting without length control, fine-tuned prompting, and the LDPE method.
Figure~\ref{fig:qa_length_comparison} presents the results of this comparison.
The corresponding ORPE results are similar to LDPE and can be found in Figure~\ref{fig:orpe_qa_length_comparison} in the Appendix.

As shown in the top-left panel of Figure \ref{fig:qa_length_comparison}, prompting without any length control results in a wide spread of response lengths, with many responses significantly deviating from the ideal length.
This highlights the need for an effective length control mechanism in question-answering systems.
The top-right panel of Figure~\ref{fig:qa_length_comparison} demonstrates the performance of fine-tuned prompting, where the model is fine-tuned with length information.
While this approach improves the alignment between the target and response lengths compared to prompting without length control, there is still a noticeable deviation from the ideal response length for many samples.
Note that these two baseline experiments were performed with Mistral only as we expect the performance to be similar for Llama.

In contrast, our LDPE length control method, shown in the bottom panels of Figure~\ref{fig:qa_length_comparison},
achieves a near-perfect alignment between the target and response lengths.
This showcases the effectiveness of the method in controlling the exact length of the generated responses in a question-answering setting.
The ORPE version of the results shows similar performance, with slightly more outliers.
Some example text responses for a range of target lengths can be seen in Table~\ref{tab:examples} in the Appendix.

\subsubsection{Summarisation}
\label{sec:summarisation}
To evaluate the performance of our content-aware length control method, we generated a dataset of summaries using the CNN/DailyMail dataset~\cite{nallapati2016abstractive}.
We selected a subset of 1000 articles from the training split and generated summaries for each article using five different system prompts, each targeting a specific summary length or style:
\begin{itemize}
	\item A one-sentence summary
	\item A one-paragraph summary
	\item A long, detailed summary not exceeding the original article length
	\item A 100-word summary
	\item A short, concise summary
\end{itemize}

The summaries were generated using the OpenAI API with the GPT-3.5-turbo-0125 model~\cite{gpt}.
For each combination of article and system prompt, we recorded the generated summary, its length, and the corresponding system prompt and article.
The resulting dataset was then used to evaluate the summarisation quality of our LDPE fine-tuned models, as well as a baseline Mistral model that was fine-tuned for prompt based length control.
Each model was prompted to generate a summary of a full article from the CNN/DailyMail dataset, and the target length of the summary was set to the length of each of the 5 GPT-3.5 summaries.
For the LDPE fine-tuned models the target length was given to the model only via the LDPE encodings, whereas for the prompt fine-tuned model the target length as was added to the prompt.
The quality of the summary was assessed by calculating the BERT scores between the generated summary and the corresponding GPT-3.5 summary.





\begin{figure}[h!]
	\centering
	\includegraphics[width=0.49\textwidth]{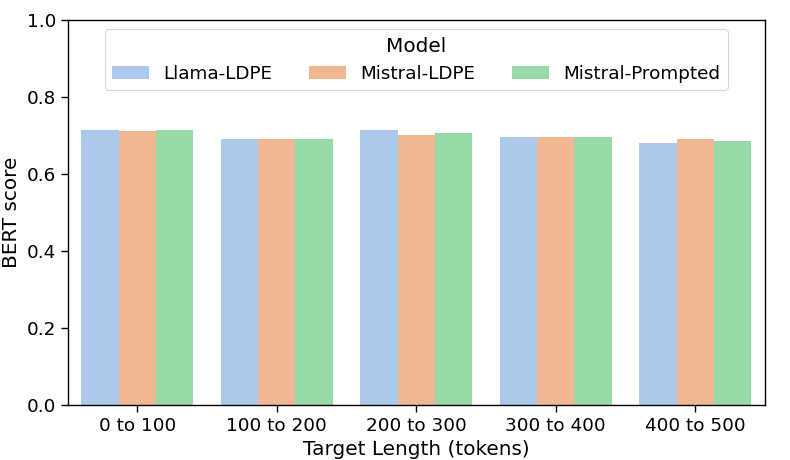}
	\includegraphics[width=0.49\textwidth]{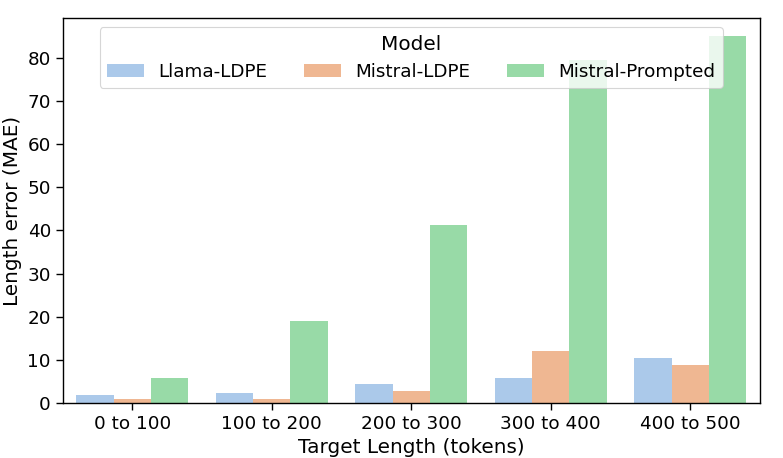}
	\caption{
		Results  for length the length controlled summarisation task.
		Models used are LDPE fine-tuned Mistral and Llama (-LDPE), as well as the
		Mistral baseline model fine-tuned for prompted based length control
		(Mistral-Prompted).
		\textbf{Left:} BERT scores between summaries and GPT-3.5 ground truth summaries.
		\textbf{Right:} Mean absolute error (MAE) between the number of tokens in model's summary and the target number of tokens.
	}
	\label{fig:bert_scores_barplot}
\end{figure}

Figure \ref{fig:bert_scores_barplot} presents the mean BERT scores for the summarisation tasks, binned into different target response length ranges.
Additional detailed summarisation quality results, including ROUGE scores, can be found in Table \ref{tab:examples} in the Appendix.
We compared both the Mistral and Llama models with the LDPE fine-tuning, as well as the baseline Mistral model fine-tuned for length control via prompting (Mistral-Prompted).
The results show that the summary quality was stable across all three models and for different response lengths.
The average BERT scores across all summaries were 0.698 for Mistral-LDPE, 0.699 for Llama-LDPE, and 0.698 the baseline Mistral-Prompted model.
The length accuracy of the summaries was significantly better for the LDPE fine-tuned models as shown in Figure \ref{fig:bert_scores_barplot}.
The mean length error between the target summary length and the models summary was 2.8 and 2.4 tokens for Mistral-LDPE and Llama-LDPE respectively.
The mean error for the Mistral-prompted was an order of magnitude larger at 24.8 tokens.
Overall the results suggest that the models are still capable of generating high-quality summaries after fine-tuning with LDPE,
and have developed the capability for fine-grained length control.



\subsubsection{Benchmarking Response Quality}
\label{sec:benchmarking}
To identify any degradation in the quality of the model's responses after fine-tuning,
we evaluated the models against a number of standard benchmarks.
We ran all benchmarks using the LM Evaluation Harness~\cite{eval-harness}.
LM Evaluation Harness evaluates models against common benchmarks using the
log-likelihood of correct or incorrect responses from the model to score the task.
The benchmarks evaluated were
the Abstraction and Reasoning Corpus (ARC)~\cite{lei2024generalized},
HellaSwag~\cite{DBLP:journals/corr/abs-1905-07830},
PIQA~\cite{DBLP:journals/corr/abs-1911-11641},
and WINOGRANDE~\cite{DBLP:journals/corr/abs-1907-10641}.


We found that the LDPE fine-tuned models had an increased reliance on the instruction tokens
used in fine-tuning. Thus instruction tokens were added into the appropriate places for each evaluation
task. Futhermore, for question answer (QA) style tasks such as ARC and PIQA the LDPE funetuned models
performed poorly unless the LDPE encodings were added to the `answer' part of the evaluation.
Likely the poor results were due to the log-likelihood of correct or incorrect response
being dominated by the additional model perplexity due to the lack of LDPE embeddings.
Therefore to evaluate the LDPE fine-tuned models on QA tasks we modified the LM Evaluation Harness to optionally
apply the LDPE countdown to the queries.
Note that this gave the model additional information about the response length, but no
information about whether a specific question response pair was correct or incorrect.



\begin{figure}[h!]
	\centering
	\includegraphics[width=0.7\textwidth]{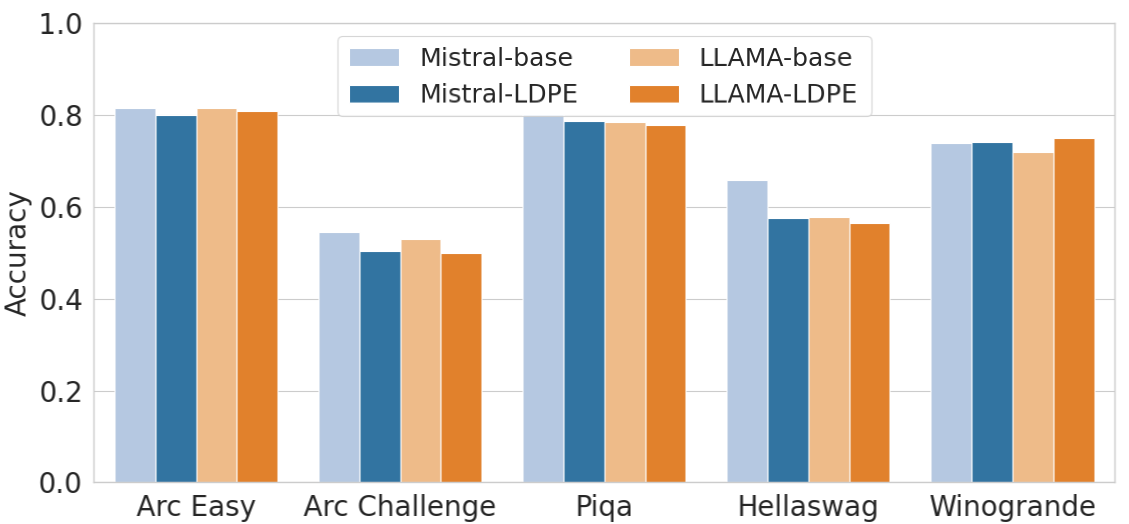}
	\caption{
		Performance of length controlled and baseline models against a range of standard benchmarks.
		All evaluations used zero-shot prompting.
		The labels Mistral and Llama correspond to Mistral-7B-Instruct and Llama-3-8B-Instruct respectively.
		The tag -LDPE means the model was fine-tuned with the LDPE. Additionally LDPE were added during the evaluation.
	}
	\label{fig:benchmarks}
\end{figure}

The results are shown in Figure~\ref{fig:benchmarks} for the LDPE fine-tuned Mistral and Llama
models (Mistral-LDPE and Llama-LDPE) as well as the base instruct versions (-base).
The results show that the LDPE fine-tuned model's performance was generally preserved after fine-tuning,
except for a reduction in Mistral's performance on the Hellaswag benchmarks.



\subsection{Max New Tokens++}
\begin{figure}[h!]
	\centering
	\includegraphics[width=0.7\textwidth]{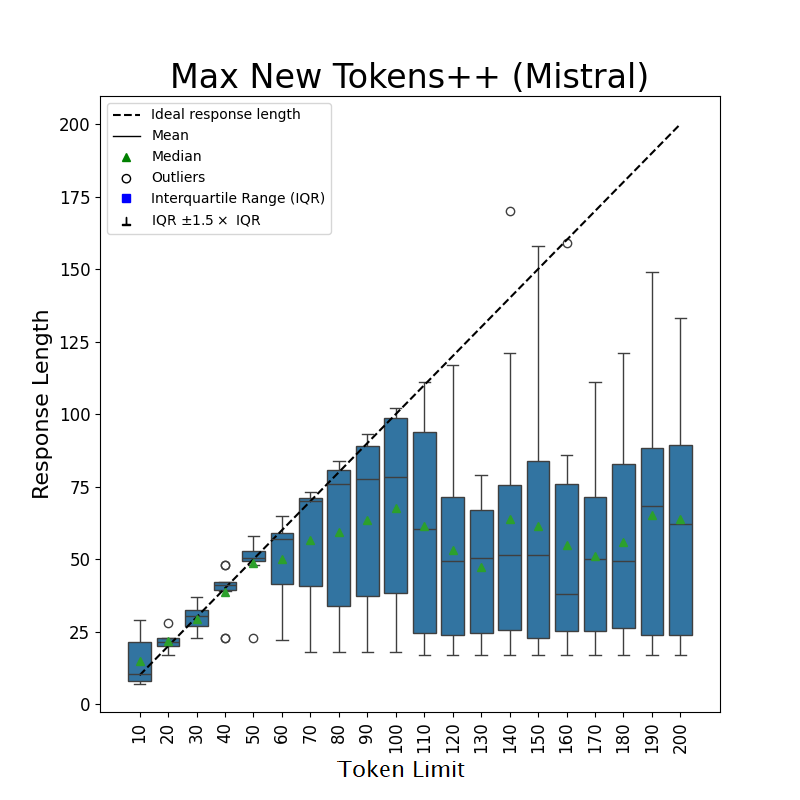}
	\caption{
		Plot of token limit vs response length for using a \textit{Max New Tokens++} fine-tuned Mistral model on a
		question-answering task.
		This technique has not been applied to Llama but we expect the performance to be fairly similar based on the similarity of the earlier results for exact length control.
	}
	\label{fig:max_new_tokens}
\end{figure}

Figure~\ref{fig:max_new_tokens} shows the comparison between the token limit and response length for a Mistral model
fine-tuned using the \textit{Max New Tokens++} method with $\sigma_0 = 0.1$ and $\sigma_\text{max} = 2048$.
For shorter token limits the responses closely follow the identity line,
while for longer token limits there is a deviation towards shorter responses.
Even though there is a shift in median away from identity the 95\% confidence interval still follows closely.
This indicates that the model still possesses a notion of \enquote{token budget},
and that it uses this token budget to smoothly terminate the longer responses as they
approach the token limit.

\section{Conclusion}
\label{sec:conclusion}
In this work we introduce a method to fine-tune decoder-only LLMs to enable precise control of the response length.
This is achieved by adding an additional reverse positional encoding to the transformer architecture.
This encoding provides the model with information about how many more tokens should be generated.
During fine-tuning, the model learns to use this countdown to structure its responses to terminate smoothly at the end of the countdown.


We investigated two different methods of applying reverse positional encodings,
Length Difference Positional Encoding (LDPE), and Offset Reverse Positional Encoding (ORPE).
The difference between the two methods is the point in the prompt-response pair at which the reverse encoding begins.
LDPE simply adds the reverse encoding to the entire prompt-response pair, whereas in ORPE the encoding is offset so that it is only applied to the model's response.
Our results demonstrate that both LDPE and ORPE effectively control the response length without degrading the underlying performance of the LLM.
There does not appear to be a significant different in performance between the two methods.

Preliminary work was also presented for the \textit{Max New Tokens++} method.
This is a way of implementing an upper bound on the number of generated tokens rather than an exact target.
The results show that this method is effective at training the model provide responses
that terminate smoothly at or before the token limit is reached.
This prevents the generation of
needlessly long answers in an attempt to meet the target length imposed by the LDPE encoding.


Overall, this work shows that the addition of a specialized positional encoding during fine-tuning enables
precise control over when the model outputs the end of sentence token, without degrading response quality.
While we focused on controlling the generation of the EOS token to terminate the response on command,
this approach could be generalised to control the generation of other tokens or potentially influence other aspects of LLM generation.


\section{Limitations}
\label{sec:limitations}


We have identified several limitations in our work that could be addressed to improve the
evaluation and effectiveness of the proposed length control method.

Firstly, we fine-tuned our models on relatively small
question-answer-focused datasets. The impact of dataset size and diversity on learning
length control remains unclear.
Secondly, our experiments were limited to the instruct versions of Mistral 7B and Llama3 8B models,
both within the 7-8B parameter range and instruction fine-tuned. While this made them suitable
for further fine-tuning with a QA dataset, it remains an open question how well the proposed
length control techniques generalise to other models.

Another limitation is that the countdown mechanism in our approach is based on tokens rather
than words or characters, the latter of which may be more relevant for user applications. Future research
could explore using LDPE to count down words or characters instead of tokens. Additionally,
incorporating the relative progress towards the termination length, rather than the absolute
number of remaining tokens, might enhance the model's ability to generalise across various
response lengths.
Finally, we utilised standard sinusoidal positional encodings in a reverse and offset order
throughout this work. These encodings, originally designed to provide information about token
ordering, might not be optimal for our specific use case where the focus is on counting down
to a termination point.

Addressing these limitations in future work would provide a more robust evaluation of the
proposed length control method and potentially enhance its applicability and performance
across different models and datasets.

\bibliographystyle{plain}  
\bibliography{references}



\begin{appendices}

\newpage
\section{Example length controlled outputs}
\label{sec:ROUGE}
Examples of length controlled outputs from the LDPE finetuned Llama model are shown in Table.~\ref{tab:examples}.
For the same prompt, the model is asked to generate responses of varied length.
This results in controlled levels of detail, while still answering the question correctly.

\begin{table}[h!]

	\resizebox{0.9\textwidth}{!}{%
		\begin{tabular}{p{0.98\linewidth} | p{0.1\linewidth} | p{0.1\linewidth}}
			\toprule
			\multicolumn{3}{p{\linewidth}}{\raggedright\textbf{Question}} \\ \midrule
			\multicolumn{3}{p{1.2\linewidth}} {\vspace{0.5em}Extract the answer to the question from the following context.
			Question: What's the expensive rock mean Context: Buddhist schools vary on the exact nature of the path to liberation, the importance and canonicity of various teachings and scriptures, and especially their respective practices.
			Buddhism denies a creator deity and posits that mundane deities such as Mahabrahma are misperceived to be a creator.
			The foundations of Buddhist tradition and practice are the Three Jewels: the Buddha, the Dharma (the teachings), and the Sangha (the community).
			Taking ""refuge in the triple gem"" has traditionally been a declaration and commitment to being on the Buddhist path, and in general distinguishes a Buddhist from a non-Buddhist.
			Other practices are Ten Meritorious Deeds including, giving charity to reduce the greediness; following ethical precepts; renouncing conventional living and becoming a monastic; the development of mindfulness and practice of meditation; cultivation of higher wisdom and discernment; study of scriptures; devotional practices; ceremonies; and in the Mahayana tradition, invocation of buddhas and bodhisattvas.
			\newline} \\ \midrule
			\textbf{Responses}                                                                                                                                                                                                                                                                                                                                                                                                                                                                                                                                                                                                                                                                                                                                                                                                                                                                                                                                                                                                                                & \textbf{Target length} & \textbf{Response length} \\ \midrule
			\vspace{0.5em}There is no answer. \newline                                                                                                                                                                                                                                                                                                                                                                                                                                                                                                                                                                                                                                                                                                                                                                                                                                                                                                                                                                                                        & 10                     & 12                       \\ \hline
			\vspace{0.5em}The context does not provide an answer to the question ""What's the expensive rock mean?"" The context is about Buddhist schools and their practices, and it does not mention anything about an expensive rock. Therefore, it is not possible to extract an answer to the question from the given context. The question seems to be unrelated to the context and requires a separate answer or information that is not provided in the given context.\newline                                                                                                                                                                                                                                                                                                                                                                                                                                                                                                                                                                       & 90                     & 89                       \\ \hline
			\vspace{0.5em}The context does not provide an answer to the question ""What's the expensive rock mean?"" The context is about Buddhist schools and their practices, including the Three Jewels, Ten Meritorious Deeds, and other practices. It also mentions the denial of a creator deity and the importance of scriptures and teachings in different Buddhist schools. However, there is no mention of an ""expensive rock"" in the context. Therefore, it is not possible to extract an answer to the question from the given context. The question seems to be unrelated to the context and requires a separate answer or information that is not provided in the given context.\newline                                                                                                                                                                                                                                                                                                                                                      & 150                    & 149                      \\ \hline
			\vspace{0.5em}The context provided does not directly answer the question about the expensive rock. However, it does provide information about the nature of the path to liberation, the importance and canonicity of various teachings and scriptures, and especially their respective practices in Buddhist schools. The context also mentions the Three Jewels, which are the Buddha, the Dharma (the teachings), and the Sangha (the community). Taking ""refuge in the triple gem"" has traditionally been a declaration and commitment to being on the Buddhist path, and in general distinguishes a Buddhist from a non-Buddhist. Therefore, the answer to the question about the expensive rock is not provided in the given context. It is important to note that the context provided is about Buddhist schools and their practices, and it does not mention anything about an expensive rock. If you have any further questions or if you would like to know more about Buddhist schools and their practices, feel free to ask.\newline & 200                    & 197                      \\
			\bottomrule
		\end{tabular}}
	\caption{
		Example generated responses using ORPE fine tuned Llama3 8B.
		Note that special tokens, instruction tokens and linebreaks have been removed from the text but not from the token count.
	}
	\label{tab:examples}
\end{table}

\newpage
\section{Summarization quality}

As explained in Section \ref{sec:summarisation}, the summarization quality of the 
LDPE finetuned models was evaluated against 1000 articles from the CNN/DailyMail dataset~\cite{nallapati2016abstractive}. 
The similarity of the summaries generated by the length controlled finetuned models were compared 
to summaries generated using the OpenAI API with the GPT-3.5-turbo-0125 model~\cite{gpt}.
In table~\ref{tab:full_summary_scores} the full results of this comparison are shown, 
including F1 ROUGE-1, ROUGE-2 and ROUGE-Lsum scores for each model and summary length range. 

\begin{table}[h!]
    \begin{tabular*}{0.9\textwidth}{p{2.7cm}p{2.5cm}p{1.0cm}p{1.0cm}p{1.0cm}p{1.3cm}@{}}
        \toprule
         Model & \raggedright{Target length (tokens)} & BERT score & R-1 & R-2 & R-Lsum \\
        \midrule
        \multirow{5}{*}{Llama-LDPE} & 0 to 100 & 0.715 & 0.522 & 0.304 & 0.420 \\
         & 100 to 200 & 0.690 & 0.497 & 0.231 & 0.331 \\
         & 200 to 300 & 0.713 & 0.554 & 0.264 & 0.380 \\
         & 300 to 400 & 0.695 & 0.544 & 0.242 & 0.403 \\
         & 400 to 500 & 0.680 & 0.531 & 0.230 & 0.403 \\
        \cline{1-6}
        \multirow{5}{*}{Mistral-LDPE} & 0 to 100 & 0.712 & 0.516 & 0.303 & 0.416 \\
         & 100 to 200 & 0.692 & 0.501 & 0.232 & 0.330 \\
         & 200 to 300 & 0.701 & 0.530 & 0.248 & 0.351 \\
         & 300 to 400 & 0.696 & 0.531 & 0.233 & 0.388 \\
         & 400 to 500 & 0.690 & 0.537 & 0.238 & 0.405 \\
        \cline{1-6}
        \multirow{5}{*}{Mistral-Prompted} & 0 to 100 & 0.714 & 0.513 & 0.301 & 0.415 \\
         & 100 to 200 & 0.690 & 0.497 & 0.230 & 0.328 \\
         & 200 to 300 & 0.706 & 0.540 & 0.260 & 0.362 \\
         & 300 to 400 & 0.696 & 0.530 & 0.233 & 0.380 \\
         & 400 to 500 & 0.686 & 0.522 & 0.233 & 0.398 \\
        \bottomrule
    \end{tabular*}
    \caption{BERT and ROUGE F1 scores between length-controlled model summaries and the GPT-3.5-turbo-0125 model~\cite{gpt}.
    Models used are LDPE fine-tuned Mistral and Llama (-LDPE), as well as the Mistral baseline model fine-tuned for prompt
    based length control (Mistral-Prompted).}
    \label{tab:full_summary_scores}
\end{table}	

\newpage
\section{Offset reverse positional encoding (ORPE) results}

We investigated the effect of offseting the reverse positional encodings so that no encoding was applied 
to the prompt. 
The logic being that in length control information should only be needed in the models response, and not in the prompt.
This is shown diagramatically by the $i_\text{ORPE}$ row in Fig \ref{fig:orpe}. 

\begin{figure*}[h!]
	\centering
	\includegraphics[width=0.49\textwidth]{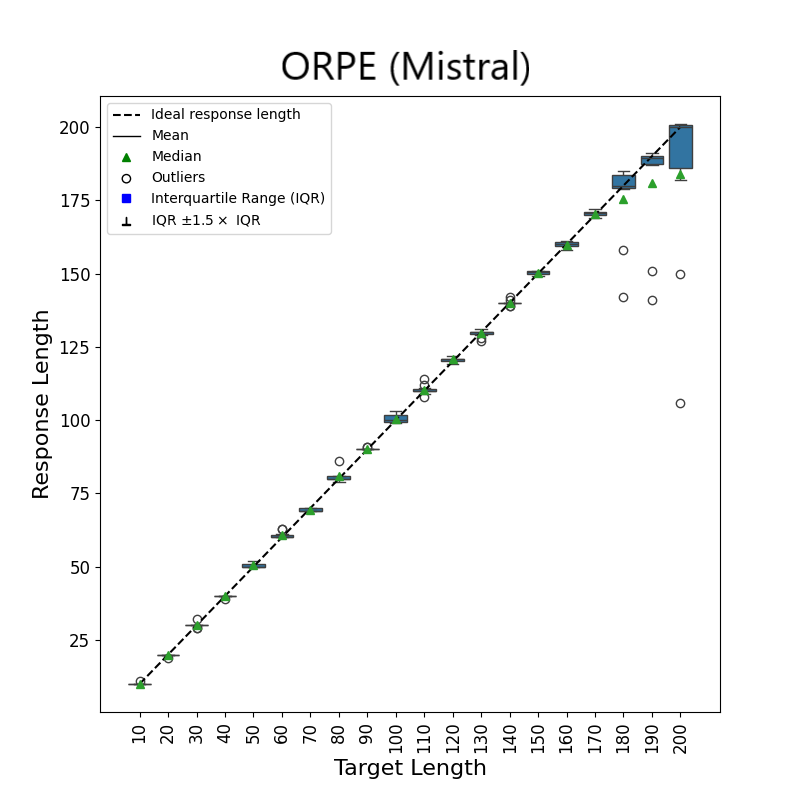}
	\includegraphics[width=0.49\textwidth]{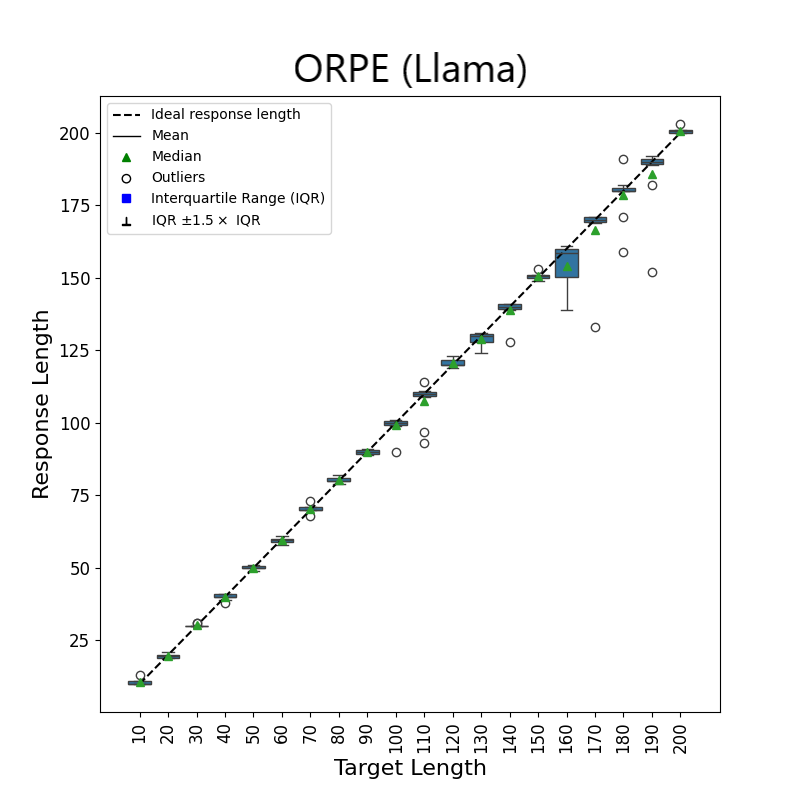}
	\caption{
		Comparison of target and response lengths for different length control approaches on a question-answering task.
		The ideal response length is indicated by the dashed line in each panel.
		\textbf{Left:} results for ORPE fine-tuned Mistral 7B model.
		\textbf{Right:} results for ORPE fine-tuned Llama3 8B model.
	}
	\label{fig:orpe_qa_length_comparison}
\end{figure*}

The results of the QA length control experiment for models fine-tuned with the ORPE version of the reverse encoding 
are shown in Fig.~\ref{fig:orpe_qa_length_comparison}.
For this experiment the same data and method was used as described in Sec.~\ref{sec:results_qa}. 
The results show effective length control, but ultimately no improvement over the slightly more straightforward LDPE technique.   

\end{appendices}

\end{document}